\documentclass{article}

\usepackage{amsmath}
\usepackage{amssymb}
\usepackage[usenames, dvipsnames]{color}
\usepackage{enumitem}
\usepackage{epsfig}
\usepackage{graphicx}
\usepackage{subcaption}
\usepackage{times}
\usepackage{xcolor}
\usepackage{microtype}

\usepackage{hyperref}

\usepackage[accepted]{icml2018}

\icmltitlerunning{GradNorm: Gradient Normalization for Adaptive Loss Balancing in Deep Multitask Networks}

\begin{document}

\twocolumn[
\icmltitle{GradNorm: Gradient Normalization for Adaptive \\Loss Balancing in Deep Multitask Networks}



\icmlsetsymbol{equal}{*}

\begin{icmlauthorlist}
\icmlauthor{Zhao Chen}{ml}
\icmlauthor{Vijay Badrinarayanan}{ml}
\icmlauthor{Chen-Yu Lee}{ml}
\icmlauthor{Andrew Rabinovich}{ml}
\end{icmlauthorlist}

\icmlaffiliation{ml}{Magic Leap, Inc}

\icmlcorrespondingauthor{Zhao Chen}{zchen@magicleap.com}

\icmlkeywords{Machine Learning, Multitask Learning, Deep Learning, ICML}

\vskip 0.3in
]



\printAffiliationsAndNotice{}  
\begin{abstract}
Deep multitask networks, in which one neural network produces multiple predictive outputs, can offer better speed and performance than their single-task counterparts but are challenging to train properly. We present a gradient normalization (GradNorm) algorithm that automatically balances training in deep multitask models by dynamically tuning gradient magnitudes. We show that for various network architectures, for both regression and classification tasks, and on both synthetic and real datasets, GradNorm improves accuracy and reduces overfitting across multiple tasks when compared to single-task networks, static baselines, and other adaptive multitask loss balancing techniques. GradNorm also matches or surpasses the performance of exhaustive grid search methods, despite only involving a single asymmetry hyperparameter $\alpha$. Thus, what was once a tedious search process that incurred exponentially more compute for each task added can now be accomplished within a few training runs, irrespective of the number of tasks. Ultimately, we will demonstrate that gradient manipulation affords us great control over the training dynamics of multitask networks and may be one of the keys to unlocking the potential of multitask learning.
\end{abstract}


\section{Introduction}\label{sec:intro}
Single-task learning in computer vision has enjoyed much success in deep learning, with many single-task models now performing at or beyond human accuracies for a wide array of tasks. However, an ultimate visual system for full scene understanding must be able to perform many diverse perceptual tasks simultaneously and efficiently, especially within the limited compute environments of embedded systems such as smartphones, wearable devices, and robots/drones. Such a system can be enabled by multitask learning, where one model shares weights across multiple tasks and makes multiple inferences in one forward pass. Such networks are not only scalable, but the shared features within these networks can induce more robust regularization and boost performance as a result. In the ideal limit, we can thus have the best of both worlds with multitask networks: more efficiency and higher performance.

In general, multitask networks are difficult to train; different tasks need to be properly balanced so network parameters converge to robust shared features that are useful across all tasks. Methods in multitask learning thus far have largely tried to find this balance by manipulating the forward pass of the network (e.g. through constructing explicit statistical relationships between features \cite{deeprelationshipnetwork} or optimizing multitask network architectures \cite{crossstitch}, etc.), but such methods ignore a key insight: \textit{task imbalances impede proper training because they manifest as imbalances between backpropagated gradients}. A task that is too dominant during training, for example, will necessarily express that dominance by inducing gradients which have relatively large magnitudes. We aim to mitigate such issues at their root by directly modifying gradient magnitudes through tuning of the multitask loss function.

In practice, the multitask loss function is often assumed to be linear in the single task losses $L_i$, $L = \sum_i w_iL_i $, where the sum runs over all $T$ tasks. In our case, we propose an adaptive method, and so $w_i$ can vary at each training step $t$: $w_i = w_i(t)$. This linear form of the loss function is convenient for implementing gradient balancing, as $w_i$ very directly and linearly couples to the backpropagated gradient magnitudes from each task. The challenge is then to find the best value for each $w_i$ at each training step $t$ that balances the contribution of each task for optimal model training. To optimize the weights $w_i(t)$ for gradient balancing, we propose a simple algorithm that penalizes the network when backpropagated gradients from any task are too large or too small. The correct balance is struck when tasks are training at similar \textit{rates}; if task $i$ is training relatively quickly, then its weight $w_i(t)$ should decrease relative to other task weights $w_j(t)|_{j\neq i}$ to allow other tasks more influence on training. Our algorithm is similar to batch normalization \cite{batchnorm} with two main differences: (1) we normalize across tasks instead of across data batches, and (2) we use rate balancing as a desired objective to inform our normalization. We will show that such gradient normalization (hereafter referred to as GradNorm) boosts network performance while significantly curtailing overfitting.

\begin{figure*}[htb!]
	\centering
	\includegraphics[width=0.85\textwidth]{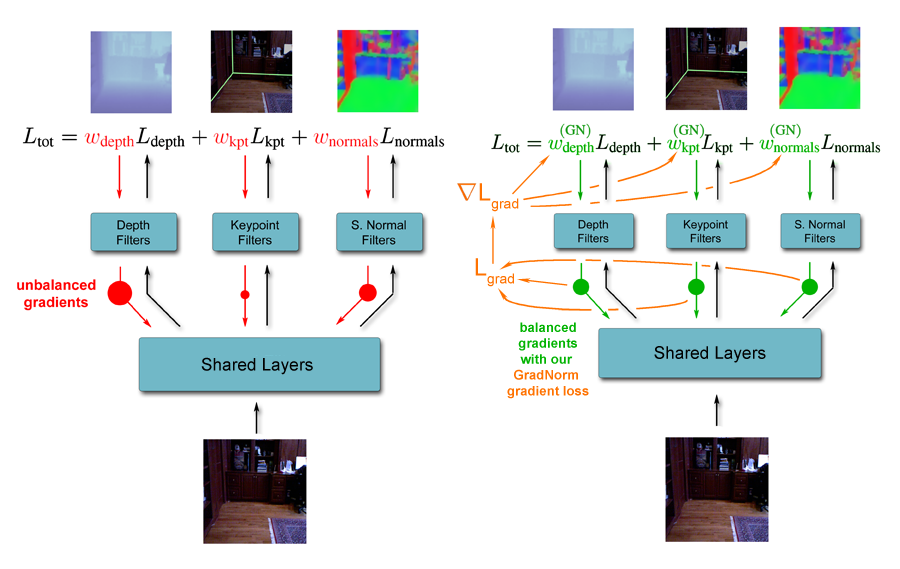}
	\caption{\textbf{Gradient Normalization.} Imbalanced gradient norms across tasks (left) result in suboptimal training within a multitask network. We implement GradNorm through computing a novel gradient loss $L_{\text{grad}}$ (right) which tunes the loss weights $w_i$ to fix such imbalances in gradient norms. We illustrate here a simplified case where such balancing results in equalized gradient norms, but in general there may be tasks that require relatively high or low gradient magnitudes for optimal training (discussed further in Section \ref{sec:methodology_0}).}
	\label{fig:diagram}
\end{figure*}

Our main contributions to multitask learning are as follows: 
\begin{enumerate}
\item An efficient algorithm for multitask loss balancing which directly tunes gradient magnitudes.
\item A method which matches or surpasses the performance of very expensive exhaustive grid search procedures, but which only requires tuning a single hyperparameter.
\item A demonstration that direct gradient interaction provides a powerful way of controlling multitask learning. 
\end{enumerate}
 
\section{Related Work}\label{sec:related_work}
Multitask learning was introduced well before the advent of deep learning \cite{caruana1998multitask, bakker2003task}, but the robust learned features within deep networks and their excellent single-task performance have spurned renewed interest. Although our primary application area is computer vision, multitask learning has applications in multiple other fields, from natural language processing \cite{nlp2collobert2008unified,nlp1hashimoto2016joint,nlp3sogaard2016deep} to speech synthesis \cite{audio2seltzer2013multi,audio1wu2015deep}, from very domain-specific applications such as traffic prediction \cite{traffichuang2014deep} to very general cross-domain work \cite{bilen2017universal}. Multitask learning has also been explored in the context of curriculum learning \cite{graves2017automated}, where subsets of tasks are subsequently trained based on local rewards; we here explore the opposite approach, where tasks are jointly trained based on global rewards such as total loss decrease.

Multitask learning is very well-suited to the field of computer vision, where making multiple robust predictions is crucial for complete scene understanding. Deep networks have been used to solve various subsets of multiple vision tasks, from 3-task networks \cite{eigen2015predicting, teichmann2016multinet} to much larger subsets as in UberNet \cite{ubernet}. Often, single computer vision problems can even be framed as multitask problems, such as in Mask R-CNN for instance segmentation \cite{maskrcnn} or YOLO-9000 for object detection \cite{yolo9000}. Particularly of note is the rich and significant body of work on finding explicit ways to exploit task relationships within a multitask model. Clustering methods have shown success beyond deep models \cite{ taskgroupingjacob2009clustered,kang2011learning}, while constructs such as deep relationship networks \cite{deeprelationshipnetwork} and cross-stich networks \cite{crossstitch} give deep networks the capacity to search for meaningful relationships between tasks and to learn which features to share between them. Work in \cite{warde2014self} and \cite{lu2016fully} use groupings amongst labels to search through possible architectures for learning. Perhaps the most relevant to the current work, \cite{kendallarxiv} uses a joint likelihood formulation to derive task weights based on the intrinsic uncertainty in each task. 
\section{The GradNorm Algorithm}\label{sec:methodology_0}
\subsection{Definitions and Preliminaries}\label{sec:methodology}
For a multitask loss function $L(t) = \sum w_i(t)L_i(t)$, we aim to learn the functions $w_i(t)$ with the following goals: (1) to place gradient norms for different tasks on a common scale through which we can reason about their relative magnitudes, and (2) to dynamically adjust gradient norms so different tasks train at similar rates. To this end, we first define the relevant quantities, first with respect to the gradients we will be manipulating.
\begin{itemize}
\item $W$: The subset of the full network weights $W\subset \mathcal{W}$ where we actually apply GradNorm. $W$ is generally chosen as the last shared layer of weights to save on compute costs\footnote{In our experiments this choice of $W$ causes GradNorm to increase training time by only $\sim 5\%$ on NYUv2.}.
\item $G_W^{(i)}(t) = ||\nabla_Ww_i(t)L_i(t)||_2$: the $L_2$ norm of the gradient of the weighted single-task loss $w_i(t)L_i(t)$ with respect to the chosen weights $W$. 
\item $\overline{G}_W(t) = E_{\text{task}}[G_W^{(i)}(t)]$: the average gradient norm across all tasks at training time $t$.
\end{itemize}
We also define various training rates for each task $i$:
\begin{itemize}
\item $\tilde{L}_i(t) = L_i(t)/L_i(0)$: the loss ratio for task $i$ at time $t$. $\tilde{L}_i(t)$ is a measure of the \textit{inverse} training rate of task $i$ (i.e. lower values of $\tilde{L}_i(t)$ correspond to a faster training rate for task $i$)\footnote{Networks in this paper all had stable initializations and $L_i(0)$ could be used directly. When $L_i(0)$ is sharply dependent on initialization, we can use a theoretical initial loss instead. E.g. for $L_i$ the CE loss across $C$ classes, we can use $L_i(0) = \log(C)$.}.
\item $r_i(t) = \tilde{L}_i(t)/E_{\text{task}}[\tilde{L}_i(t)]$: the relative \textit{inverse} training rate of task $i$.
\end{itemize}
With the above definitions in place, we now complete our description of the GradNorm algorithm. 
\subsection{Balancing Gradients with GradNorm}\label{sec:methodology_2}
As stated in Section \ref{sec:methodology}, GradNorm should establish a common scale for gradient magnitudes, and also should balance training rates of different tasks. The common scale for gradients is most naturally the average gradient norm, $\overline{G}_W(t)$, which establishes a baseline at each timestep $t$ by which we can determine relative gradient sizes. The relative inverse training rate of task $i$, $r_i(t)$, can be used to rate balance our gradients. Concretely, the higher the value of $r_i(t)$, the higher the gradient magnitudes should be for task $i$ in order to encourage the task to train more quickly. Therefore, our desired gradient norm for each task $i$ is simply:

\begin{equation}\label{eq:gradvalue}
G_W^{(i)}(t)  \mapsto \overline{G}_W(t)\times [r_i(t)]^{\alpha},
\end{equation}

where $\alpha$ is an additional hyperparameter. $\alpha$ sets the strength of the restoring force which pulls tasks back to a common training rate. In cases where tasks are very different in their complexity, leading to dramatically different learning dynamics between tasks, a higher value of $\alpha$ should be used to enforce stronger training rate balancing. When tasks are more symmetric (e.g. the synthetic examples in Section \ref{sec:toy_example}), a lower value of $\alpha$ is appropriate. Note that $\alpha=0$ will always try to pin the norms of backpropagated gradients from each task to be equal at $W$. See Section \ref{sec:alpha_tuning} for more details on the effects of tuning $\alpha$. 

Equation \ref{eq:gradvalue} gives a target for each task $i$'s gradient norms, and we update our loss weights $w_i(t)$ to move gradient norms towards this target for each task. GradNorm is then implemented as an $L_1$ loss function $L_{\text{grad}}$ between the actual and target gradient norms at each timestep for each task, summed over all tasks:
\begin{equation}
 L_{\text{grad}}(t; w_i(t)) = \sum_i\biggr\rvert G_W^{(i)}(t) - \overline{G}_W(t)\times [r_i(t)]^{\alpha}\biggr\rvert_1
\end{equation}
where the summation runs through all $T$ tasks. When differentiating this loss $L_{\text{grad}}$, we treat the target gradient norm $\overline{G}_W(t)\times [r_i(t)]^{\alpha}$ as a fixed constant to prevent loss weights $w_i(t)$ from spuriously drifting towards zero. $L_{\text{grad}}$ is then differentiated \textit{only with respect to the $w_i$}, as the $w_i(t)$ directly control gradient magnitudes per task. The computed gradients $\nabla_{w_i} L_{\text{grad}}$ are then applied via standard update rules to update each $w_i$ (as shown in Figure \ref{fig:diagram}). The full GradNorm algorithm is summarized in Algorithm \ref{alg:gradnorm}. Note that after every update step, we also renormalize the weights $w_i(t)$ so that $\sum_i w_i(t) = T$ in order to decouple gradient normalization from the global learning rate. 

\begin{algorithm}[tb]
   \caption{Training with GradNorm}
   \label{alg:gradnorm}
\begin{algorithmic}
   \STATE Initialize $w_i(0)=1$ $\forall i$
   \STATE Initialize network weights $\mathcal{W}$
   \STATE Pick value for $\alpha> 0$ and pick the weights $W$ (usually the \indent \hspace{0.4cm}final layer of weights which are shared between tasks)
   \FOR{$t=0$ {\bfseries to} $max\_train\_steps$}
   \STATE {\bfseries Input} batch $x_i$ to compute $L_i(t)$ $\forall i$ and \\ \indent \hspace{0.4cm} $L(t) = \sum_i w_i(t)L_i(t)$ [standard forward pass]
   \STATE Compute $G_W^{(i)}(t)$ and $r_i(t)$ $\forall i$
   \STATE Compute $\overline{G}_W(t)$ by averaging the $G_W^{(i)}(t)$
   \STATE Compute $L_{\text{grad}}= \sum_i\rvert G_W^{(i)}(t) - \overline{G}_W(t)\times [r_i(t)]^{\alpha}\rvert_1$
   \STATE Compute GradNorm gradients $\nabla_{w_i} L_{\text{grad}}$, keeping \\ \indent \hspace{0.4cm}targets $\overline{G}_W(t)\times [r_i(t)]^{\alpha}$ constant
   \STATE Compute standard gradients $\nabla_{\mathcal{W}} L(t)$
   \STATE Update $w_i(t) \mapsto w_i(t+1)$ using $\nabla_{w_i} L_{\text{grad}}$
   \STATE Update $\mathcal{W}(t) \mapsto \mathcal{W}(t+1)$ using $\nabla_{\mathcal{W}} L(t)$ [standard \\ \indent \hspace{0.4cm}backward pass]
   \STATE Renormalize $w_i(t+1)$ so that $\sum_iw_i(t+1) = T$
   \ENDFOR
\end{algorithmic}
\end{algorithm}

\section{A Toy Example}\label{sec:toy_example}
To illustrate GradNorm on a simple, interpretable system, we construct a common scenario for multitask networks: training tasks which have similar loss functions but different loss scales. In such situations, if we na\"{i}vely pick $w_i(t)=1$ for all loss weights $w_i(t)$, the network training will be dominated by tasks with larger loss scales that backpropagate larger gradients. We will demonstrate that GradNorm overcomes this issue.

Consider $T$ regression tasks trained using standard squared loss onto the functions \begin{equation} f_i(\textbf{x}) = \sigma_i \tanh((B+\epsilon_i)\textbf{x}),\end{equation} where $\tanh(\cdot)$ acts element-wise. Inputs are dimension 250 and outputs dimension 100, while $B$ and $\epsilon_i$ are constant matrices with their elements generated IID from $\mathcal{N}(0,10)$ and $\mathcal{N}(0,3.5)$, respectively. Each task therefore shares information in $B$ but also contains task-specific information $\epsilon_i$. The $\sigma_i$ are the key parameters within this problem; they are fixed scalars which set the scales of the outputs $f_i$. A higher scale for $f_i$ induces a higher expected value of squared loss for that task. Such tasks are harder to learn due to the higher variances in their response values, but they also backpropagate larger gradients. This scenario generally leads to suboptimal training dynamics when the higher $\sigma_i$ tasks dominate the training across all tasks.

To train our toy models, we use a 4-layer fully-connected ReLU-activated network with 100 neurons per layer as a common trunk. A final affine transformation layer gives $T$ final predictions (corresponding to $T$ different tasks). To ensure valid analysis, we only compare models initialized to the same random values and fed data generated from the same fixed random seed. The asymmetry $\alpha$ is set low to 0.12 for these experiments, as the output functions $f_i$ are all of the same functional form and thus we expect the asymmetry between tasks to be minimal. 

In these toy problems, we measure the \textit{task-normalized} test-time loss to judge test-time performance, which is the sum of the test loss ratios for each task, $\sum_i L_i(t)/L_i(0)$. We do this because a simple sum of losses is an inadequate performance metric for multitask networks when different loss scales exist; higher loss scale tasks will factor disproportionately highly in the loss. There unfortunately exists no general single scalar which gives a meaningful measure of multitask performance in all scenarios, but our toy problem was specifically designed with tasks which are statistically identical except for their loss scales $\sigma_i$. There is therefore a clear measure of overall network performance, which is the sum of losses normalized by each task's variance $\sigma_i^2$ - equivalent (up to a scaling factor) to the sum of loss ratios.

\begin{figure*}[htb!]
	\centering
	\begin{subfigure}[h]{0.21\linewidth}
		\includegraphics[width=\textwidth]{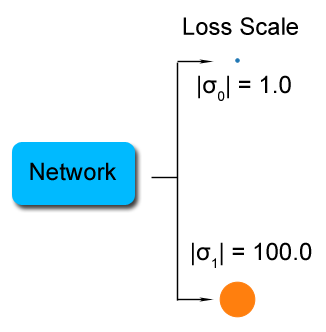}       						   
	\end{subfigure}
	~
	\begin{subfigure}[h]{0.29\linewidth}
		\includegraphics[width=\textwidth]{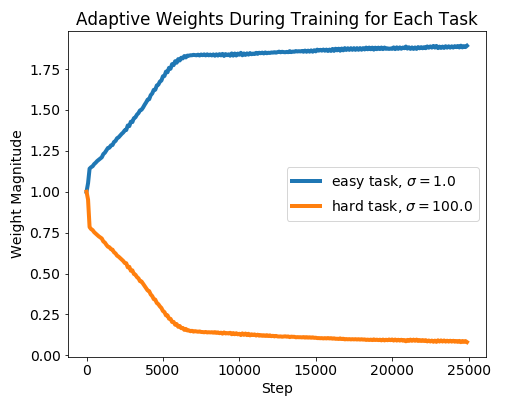}
	\end{subfigure}
    ~
	\begin{subfigure}[h]{0.29\linewidth}
		\includegraphics[width=\textwidth]{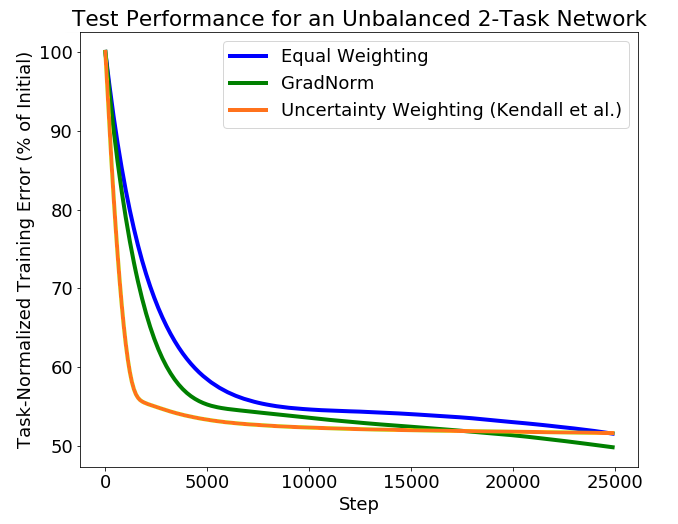}       						   
	\end{subfigure}
	\\
	\begin{subfigure}[h]{0.21\linewidth}
		\includegraphics[width=\textwidth]{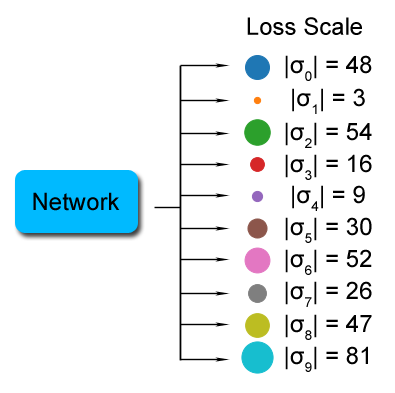}       						   
	\end{subfigure}
    ~
	\begin{subfigure}[h]{0.29\linewidth}
		\includegraphics[width=\textwidth]{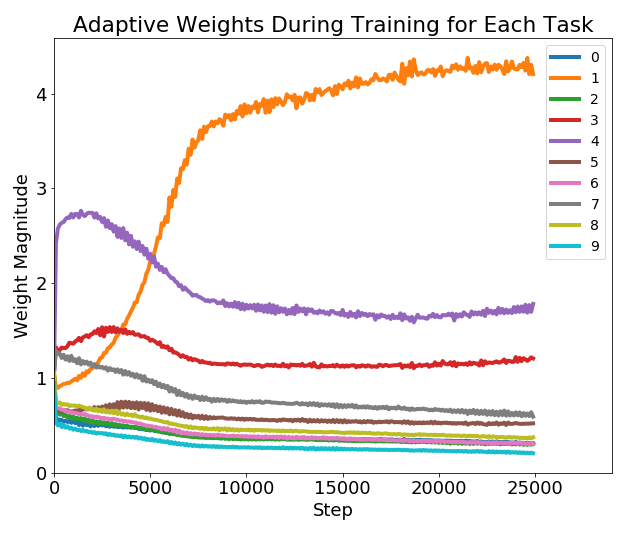}
	\end{subfigure}
    ~
	\begin{subfigure}[h]{0.29\linewidth}
		\includegraphics[width=\textwidth]{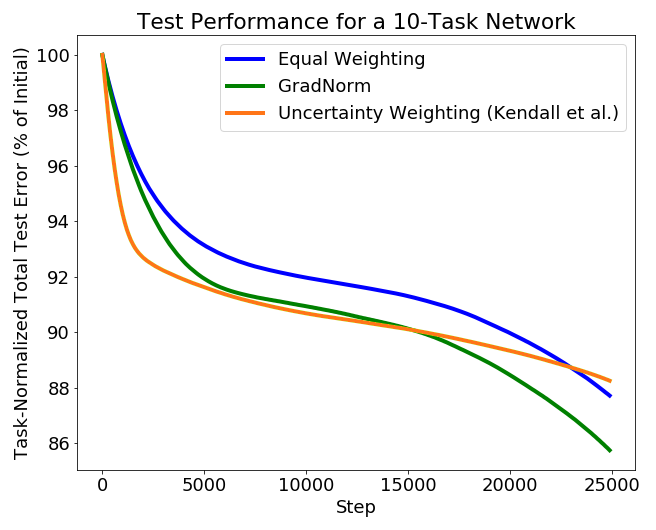}       						   
	\end{subfigure}
	\caption{\textbf{Gradient Normalization on a toy 2-task (top) and 10-task (bottom) system.} Diagrams of the network structure with loss scales are on the left, traces of $w_i(t)$ during training in the middle, and task-normalized test loss curves on the right. $\alpha=0.12$ for all runs.}
	\label{fig:toy_problem}
\end{figure*}

For $T=2$, we choose the values $(\sigma_0,\sigma_1) = (1.0,100.0)$ and show the results of training in the top panels of Figure \ref{fig:toy_problem}. If we train with equal weights $w_i = 1$, task 1 suppresses task 0 from learning due to task 1's higher loss scale. However, gradient normalization increases $w_0(t)$ to counteract the larger gradients coming from $T_1$, and the improved task balance results in better test-time performance.

The possible benefits of gradient normalization become even clearer when the number of tasks increases. For $T=10$, we sample the $\sigma_i$ from a wide normal distribution and plot the results in the bottom panels of Figure \ref{fig:toy_problem}. GradNorm significantly improves test time performance over na\"{i}vely weighting each task the same. Similarly to the $T=2$ case, for $T=10$ the $w_i(t)$ grow larger for smaller $\sigma_i$ tasks.

For both $T=2$ and $T=10$, GradNorm is more stable and outperforms the uncertainty weighting proposed by \cite{kendallarxiv}. Uncertainty weighting, which enforces that $w_i(t) \sim 1/L_i(t)$, tends to grow the weights $w_i(t)$ too large and too quickly as the loss for each task drops. Although such networks train quickly at the onset, the training soon deteriorates. This issue is largely caused by the fact that uncertainty weighting allows $w_i(t)$ to change without constraint (compared to GradNorm which ensures $\sum w_i(t) = T$ always), which pushes the global learning rate up rapidly as the network trains.

The traces for each $w_i(t)$ during a single GradNorm run are observed to be stable and convergent. In Section \ref{sec:gridsearch} we will see how the time-averaged weights $E_t[w_i(t)]$ lie close to the optimal static weights, suggesting GradNorm can greatly simplify the tedious grid search procedure.

\section{Application to a Large Real-World Dataset}\label{sec:dataset}
We use two variants of NYUv2 \cite{nyuv2} as our main datasets. Please refer to the Supplementary Materials for additional results on a 9-task facial landmark dataset found in \cite{zhang2014facial}. The standard NYUv2 dataset carries depth, surface normals, and semantic segmentation labels (clustered into 13 distinct classes) for a variety of indoor scenes in different room types (bathrooms, living rooms, studies, etc.). NYUv2 is relatively small (795 training, 654 test images), but contains both regression and classification labels, making it a good choice to test the robustness of GradNorm across various tasks. 

We augment the standard NYUv2 depth dataset with flips and additional frames from each video, resulting in 90,000 images complete with pixel-wise depth, surface normals, and room keypoint labels (segmentation labels are, unfortunately, not available for these additional frames). Keypoint labels are professionally annotated by humans, while surface normals are generated algorithmically. The full dataset is then split by scene for a 90/10 train/test split. See Figure \ref{fig:inference} for examples. We will generally refer to these two datasets as NYUv2+seg and NYUv2+kpts, respectively. 

All inputs are downsampled to 320 x 320 pixels and outputs to 80 x 80 pixels. We use these resolutions following \cite{lee2017roomnet}, which represents the state-of-the-art in room keypoint prediction and from which we also derive our VGG-style model architecture. These resolutions also allow us to keep models relatively slim while not compromising semantic complexity in the ground truth output maps.
\subsection{Model and General Training Characteristics}\label{sec:model}
We try two different models: (1) a SegNet \cite{badrinarayanan2015segnet, lee2017roomnet} network with a symmetric VGG16 \cite{vgg} encoder/decoder, and (2) an FCN \cite{fcn} network with a modified ResNet-50 \cite{resnet} encoder and shallow ResNet decoder. The VGG SegNet reuses maxpool indices to perform upsampling, while the ResNet FCN learns all upsampling filters. The ResNet architecture is further thinned (both in its filters and activations) to contrast with the heavier, more complex VGG SegNet: stride-2 layers are moved earlier and all 2048-filter layers are replaced by 1024-filter layers. Ultimately, the VGG SegNet has 29M parameters versus 15M for the thin ResNet. All model parameters are shared amongst all tasks until the final layer. Although we will focus on the VGG SegNet in our more in-depth analysis, by designing and testing on two extremely different network topologies we will further demonstrate GradNorm's robustness to the choice of base architecture. 

We use standard pixel-wise loss functions for each task: cross entropy for segmentation, squared loss for depth, and cosine similarity for normals. As in \cite{lee2017roomnet}, for room layout we generate Gaussian heatmaps for each of 48 room keypoint types and predict these heatmaps with a pixel-wise squared loss. Note that all regression tasks are quadratic losses (our surface normal prediction uses a cosine loss which is quadratic to leading order), allowing us to use $r_i(t)$ for each task $i$ as a direct proxy for each task's relative inverse training rate.   

All runs are trained at a batch size of 24 across 4 Titan X GTX 12GB GPUs and run at 30fps on a single GPU at inference. All NYUv2 runs begin with a learning rate of 2e-5. NYUv2+kpts runs last 80000 steps with a learning rate decay of 0.2 every 25000 steps. NYUv2+seg runs last 20000 steps with a learning rate decay of 0.2 every 6000 steps. Updating $w_i(t)$ is performed at a learning rate of 0.025 for both GradNorm and the uncertainty weighting (\cite{kendallarxiv}) baseline. All optimizers are Adam, although we find that GradNorm is insensitive to the optimizer chosen. We implement GradNorm using TensorFlow v1.2.1.
\subsection{Main Results on NYUv2}\label{sec:performance}

\begin{table}
\caption{Test error, NYUv2+seg for GradNorm and various baselines. Lower values are better. Best performance for each task is bolded, with second-best underlined.}\vspace{0.2cm}
\footnotesize
\centering
\begin{tabular}{c c c c}
\hline
\textbf{Model and}  & \textbf{Depth}  & \textbf{Seg.} & \textbf{Normals} \\
\textbf{Weighting}&\textbf{RMS Err.}&\textbf{Err.}&\textbf{Err.}\\
\textbf{Method}&\textbf{(m)}&\textbf{(100-IoU)}&\textbf{(1-$|$cos$|$)} \\
\hline
\underline{VGG Backbone}\\
Depth Only & 1.038 & - & -\\
Seg. Only & - & 70.0 & -\\
Normals Only & - & - & \textbf{0.169}\\
Equal Weights & 0.944& 70.1&0.192\\
GradNorm Static& \underline{0.939}& \textbf{67.5} & \underline{0.171} \\
GradNorm $\alpha=1.5$ &\textbf{0.925}&\underline{67.8}& 0.174\\
\end{tabular}

\label{table:testaccseg}
\end{table}
\begin{figure}[htb!]
	\centering
	\includegraphics[width=0.5\textwidth]{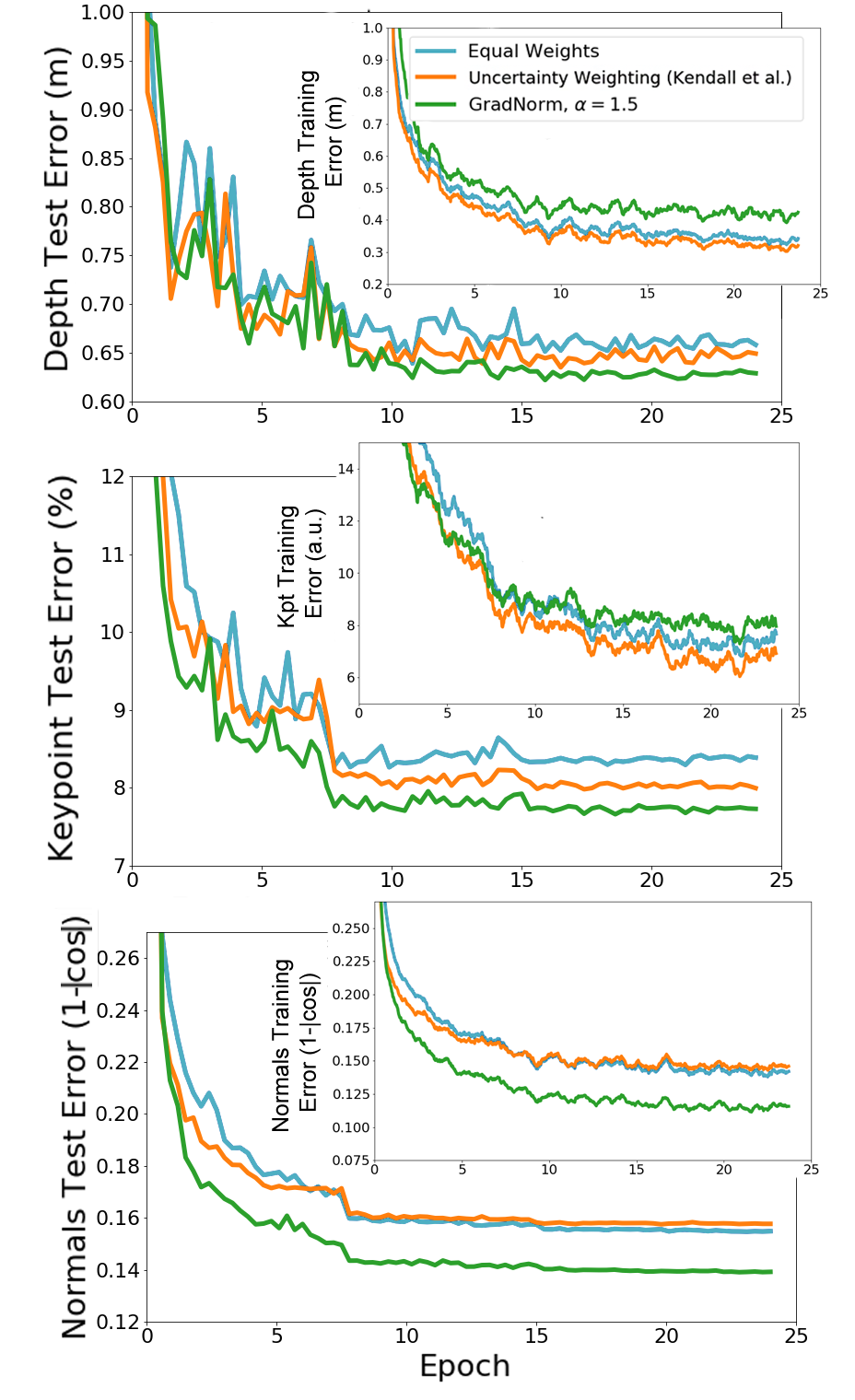}
	\caption{\textbf{Test and training loss curves for all tasks in NYUv2+kpts, VGG16 backbone}. GradNorm versus an equal weights baseline and uncertainty weighting \cite{kendallarxiv}.}
	\label{fig:loss_curves}
\end{figure}

In Table \ref{table:testaccseg} we display the performance of GradNorm on the NYUv2+seg dataset. We see that GradNorm $\alpha=1.5$ improves the performance of all three tasks with respect to the equal-weights baseline (where $w_i(t) = 1$ for all $t$,$i$), and either surpasses or matches (within statistical noise) the best performance of single networks for each task. The GradNorm Static network uses static weights derived from a GradNorm network by calculating the time-averaged weights $E_{t}[w_i(t)]$ for each task during a GradNorm training run, and retraining a network with weights fixed to those values. GradNorm thus can also be used to extract good values for static weights. We pursue this idea further in Section \ref{sec:gridsearch} and show that these weights lie very close to the optimal weights extracted from exhaustive grid search.

\begin{table}[h]
\caption{Test error, NYUv2+kpts for GradNorm and various baselines. Lower values are better. Best performance for each task is bolded, with second-best underlined.}\vspace{0.2cm}
\footnotesize
\centering
\begin{tabular}{c c c c}
\hline
\textbf{Model and}  & \textbf{Depth}  & \textbf{Kpt.} & \textbf{Normals} \\
\textbf{Weighting}&\textbf{RMS Err.}&\textbf{Err.}&\textbf{Err.}\\
\textbf{Method}&\textbf{(m)}&\textbf{(\%)}&\textbf{(1-$|$cos$|$)}\\
\hline
\underline{ResNet Backbone}\\
Depth Only & 0.725 & - & -\\
Kpt Only & - & 7.90 & -\\
Normals Only & - & - & \textbf{0.155}\\
Equal Weights& 0.697 & 7.80 & 0.172\\
\cite{kendallarxiv}& 0.702 & 7.96 & 0.182
\\
GradNorm Static& \underline{0.695}&\underline{7.63}&\underline{0.156}\\
GradNorm $\alpha=1.5$ &\textbf{0.663}&\textbf{7.32}& \textbf{0.155}\\
\hline
\underline{VGG Backbone}\\
Depth Only & 0.689 & - & -\\
Keypoint Only & - & 8.39 & -\\
Normals Only & - & - & 0.142\\
Equal Weights& 0.658 & 8.39 & 0.155\\
\cite{kendallarxiv}& 0.649 & 8.00 & 0.158
\\
GradNorm Static &\underline{0.638}& \textbf{7.69}& \textbf{0.137}\\
GradNorm $\alpha=1.5$ &\textbf{0.629}&\underline{7.73}& \underline{0.139}\\
\end{tabular}
\label{table:testacc}
\end{table}

\begin{figure}[htb!]
	\centering
	\includegraphics[width=0.45\textwidth]{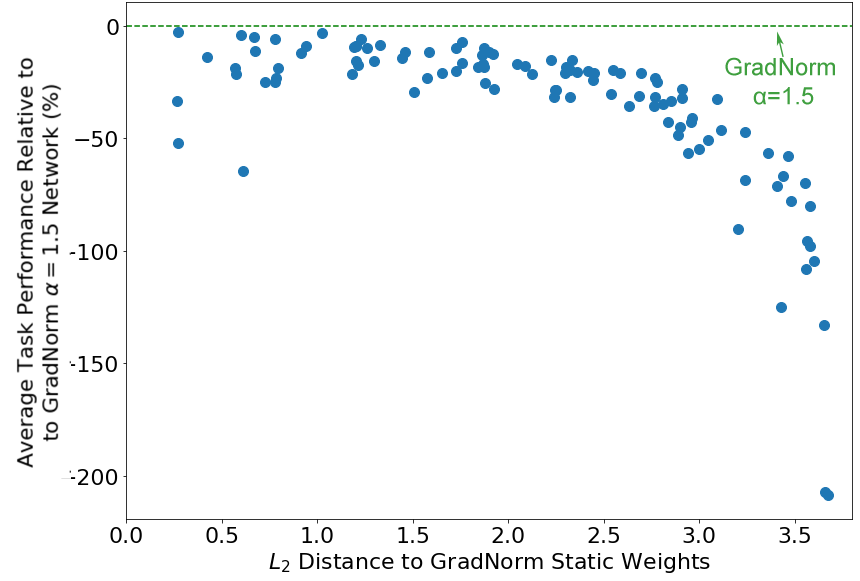}
	\caption{\textbf{Gridsearch performance for random task weights vs GradNorm, NYUv2+kpts}. Average change in performance across three tasks for a static multitask network with weights $w^{\text{static}}_i$, plotted against the $L_2$ distance between $w^{\text{static}}_i$ and a set of static weights derived from a GradNorm network, $E_t[w_i(t)]$. A reference line at zero performance change is provided for convenience. All comparisons are made at 15000 steps of training.}
	\label{fig:gridsearch}
\end{figure}

To show how GradNorm can perform in the presence of a larger dataset, we also perform extensive experiments on the NYUv2+kpts dataset, which is augmented to a factor of 50x more data. The results are shown in Table \ref{table:testacc}. As with the NYUv2+seg runs, GradNorm networks outperform other multitask methods, and either matches (within noise) or surpasses the performance of single-task networks. 

Figure \ref{fig:loss_curves} shows test and training loss curves for GradNorm ($\alpha=1.5$) and baselines on the larger NYUv2+kpts dataset for our VGG SegNet models. GradNorm improves test-time depth error by $\sim 5\%$, despite converging to a much higher training loss. GradNorm achieves this by aggressively rate balancing the network (enforced by a high asymmetry $\alpha=1.5$), and ultimately suppresses the depth weight $w_{\text{depth}}(t)$ to lower than 0.10 (see Section \ref{sec:alpha_tuning} for more details). The same trend exists for keypoint regression, and is a clear signal of network regularization. In contrast, uncertainty weighting \cite{kendallarxiv} always moves test and training error in the same direction, and thus is not a good regularizer. Only results for the VGG SegNet are shown here, but the Thin ResNet FCN produces consistent results. 
\begin{figure}[htb!]
	\centering
	\includegraphics[width=0.45\textwidth]{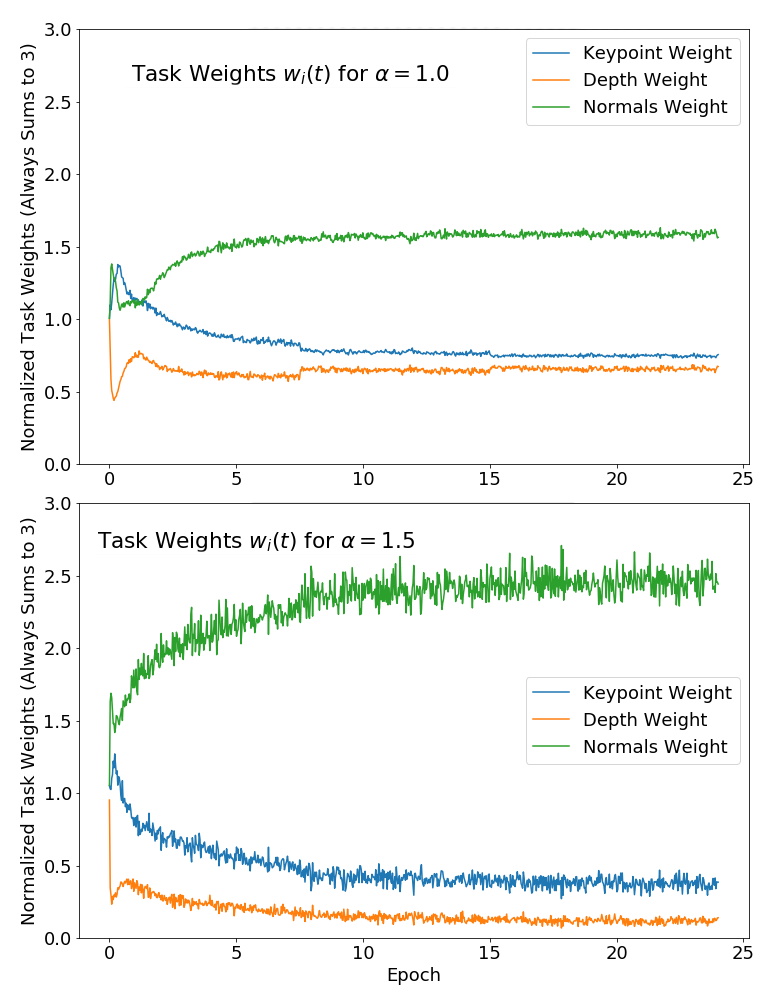}
	\caption{\textbf{Weights $w_i(t)$ during training, NYUv2+kpts.} Traces of how the task weights $w_i(t)$ change during training for two different values of $\alpha$. A larger value of $\alpha$ pushes weights farther apart, leading to less symmetry between tasks.}
	\label{fig:tuning_alpha}
\end{figure}
\subsection{Gradient Normalization Finds Optimal Grid-Search Weights in One Pass}\label{sec:gridsearch} For our VGG SegNet, we train 100 networks from scratch with random task weights on NYUv2+kpts. Weights are sampled from a uniform distribution and renormalized to sum to $T=3$. For computational efficiency, we only train for 15000 iterations out of the normal 80000, and then compare the performance of that network to our GradNorm $\alpha=1.5$ VGG SegNet network at the same 15000 steps. The results are shown in Figure \ref{fig:gridsearch}.

Even after 100 networks trained, grid search still falls short of our GradNorm network. Even more remarkably, there is a strong, negative correlation between network performance and task weight distance to our time-averaged GradNorm weights $E_t[w_i(t)]$. At an $L_2$ distance of $\sim 3$, grid search networks on average have almost double the errors per task compared to our GradNorm network. GradNorm has therefore \textit{found the optimal grid search weights in one single training run}. 


\subsection{Effects of tuning the asymmetry $\alpha$} \label{sec:alpha_tuning}
The only hyperparameter in our algorithm is the asymmetry $\alpha$. The optimal value of $\alpha$ for NYUv2 lies near $\alpha=1.5$, while in the highly symmetric toy example in Section \ref{sec:toy_example} we used $\alpha = 0.12$. This observation reinforces our characterization of $\alpha$ as an asymmetry parameter.

Tuning $\alpha$ leads to performance gains, but we found that for NYUv2, almost any value of $0<\alpha<3$ will improve network performance over an equal weights baseline (see Supplementary for details). Figure $\ref{fig:tuning_alpha}$ shows that higher values of $\alpha$ tend to push the weights $w_i(t)$ further apart, which more aggressively reduces the influence of tasks which overfit or learn too quickly (in our case, depth). Remarkably, at $\alpha=1.75$ (not shown) $w_{\text{depth}}(t)$ is suppressed to below 0.02 at no detriment to network performance on the depth task. 
 
\begin{figure*}[htb!]
	\centering
	\includegraphics[width=0.88\textwidth]{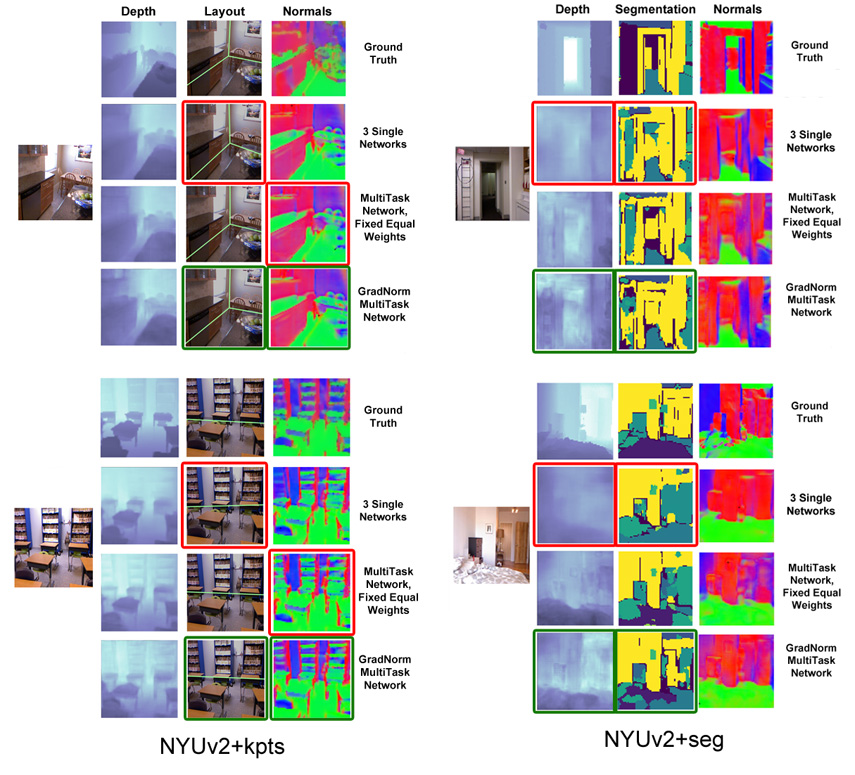}
	\caption{\textbf{Visualizations at inference time.} NYUv2+kpts outputs are shown on the left, while NYUv2+seg outputs are shown on the right. Visualizations shown were generated from random test set images. Some improvements are incremental, but red frames are drawn around predictions that are visually more clearly improved by GradNorm. For NYUv2+kpts outputs GradNorm shows improvement over the equal weights network in normals prediction and over single networks in keypoint prediction. For NYUv2+seg there is an improvement over single networks in depth and segmentation accuracy. These are consistent with the numbers reported in Tables \ref{table:testaccseg} and \ref{table:testacc}.}
	\label{fig:inference}
\end{figure*}
\subsection{Qualitative Results}

Figure \ref{fig:inference} shows visualizations of the VGG SegNet outputs on test set images along with the ground truth, for both the NYUv2+seg and NYUv2+kpts datasets. Ground truth labels are juxtaposed with outputs from the equal weights network, 3 single networks, and our best GradNorm network. Some improvements are incremental, but GradNorm produces superior visual results in tasks for which there are significant quantitative improvements in Tables \ref{table:testaccseg} and \ref{table:testacc}.

\section{Conclusions}
We introduced GradNorm, an efficient algorithm for tuning loss weights in a multi-task learning setting based on balancing the training rates of different tasks. We demonstrated on both synthetic and real datasets that GradNorm improves multitask test-time performance in a variety of scenarios, and can accommodate various levels of asymmetry amongst the different tasks through the hyperparameter $\alpha$. Our empirical results indicate that GradNorm offers superior performance over state-of-the-art multitask adaptive weighting methods and can match or surpass the performance of exhaustive grid search while being significantly less time-intensive. 

Looking ahead, algorithms such as GradNorm may have applications beyond multitask learning. We hope to extend the GradNorm approach to work with class-balancing and sequence-to-sequence models, all situations where problems with conflicting gradient signals can degrade model performance. We thus believe that our work not only provides a robust new algorithm for multitask learning, but also reinforces the powerful idea that gradient tuning is fundamental for training large, effective models on complex tasks.  
\clearpage
\bibliographystyle{icml2018}
\bibliography{ref}
\clearpage
\section{GradNorm: Gradient Normalization for Adaptive Loss Balancing in Deep Multitask Networks: Supplementary Materials}
\subsection{Performance Gains Versus $\alpha$}

The $\alpha$ asymmetry hyperparameter, we argued, allows us to accommodate for various different priors on the symmetry between tasks. A low value of $\alpha$ results in gradient norms which are of similar magnitude across tasks, ensuring that each task has approximately equal impact on the training dynamics throughout training. A high value of $\alpha$ will penalize tasks whose losses drop too quickly, instead placing more weight on tasks whose losses are dropping more slowly.

\begin{figure}[htb!]
	\centering
	\includegraphics[width=0.5\textwidth]{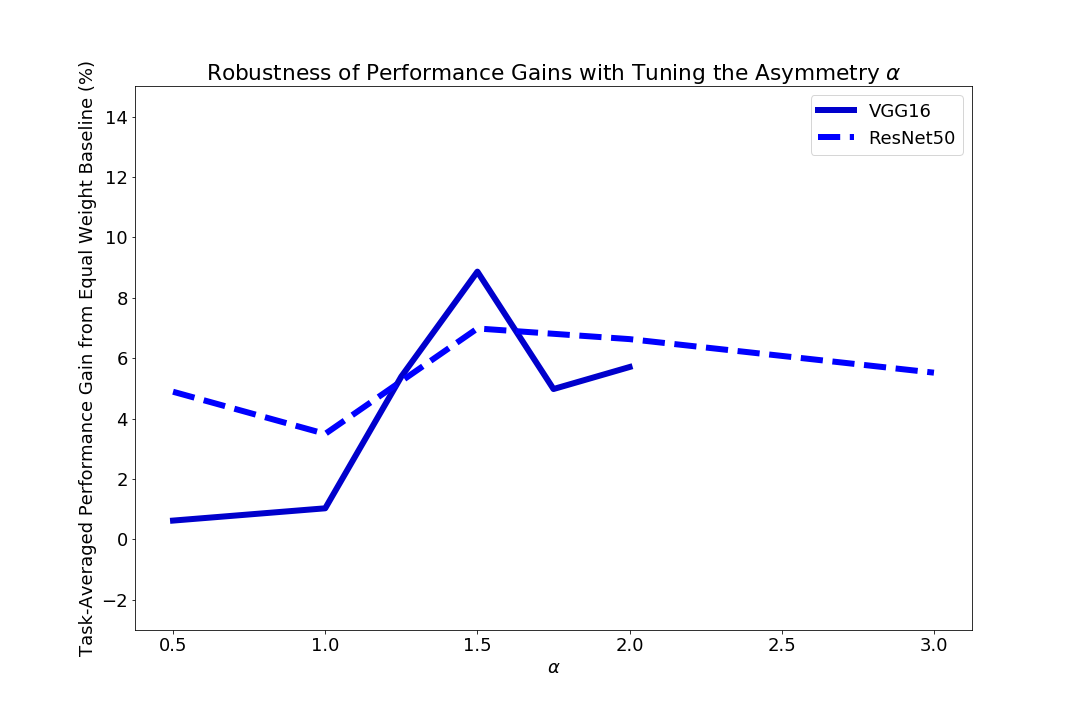}
	\caption{\textbf{Performance gains on NYUv2+kpts for various settings of $\alpha$}. For various values of $\alpha$, we plot the average performance gain (defined as the mean of the percent change in the test loss compared to the equal weights baseline across all tasks) on NYUv2+kpts. We show results for both the VGG16 backbone (solid line) and the ResNet50 backbone (dotted line). We show performance gains at all values of $\alpha$ tested, although gains appear to peak around $\alpha=1.5$. No points past $\alpha>2$ are shown for the VGG16 backbone as GradNorm weights are unstable past this point for this particular architectural backbone.}
	\label{fig:perf_gains_alpha}
\end{figure}

For our NYUv2 experiments, we chose $\alpha=1.5$ as our optimal value for $\alpha$, and in Section 5.4 we touched upon how increasing $\alpha$ pushes the task weights $w_i(t)$ farther apart. It is interesting to note, however, that we achieve overall gains in performance for almost all positive values of $\alpha$ for which GradNorm is numerically stable\footnote{At large positive values of $\alpha$, which in the NYUv2 case corresponded to $\alpha \geq 3$, some weights were pushed too close to zero and GradNorm updates became unstable.}. These results are summarized in Figure \ref{fig:perf_gains_alpha}. 

We see from Figure \ref{fig:perf_gains_alpha} that we achieve performance gains at almost all values of $\alpha$. However, for NYUv2+kpts in particular, these performance gains seem to be peaked at $\alpha=1.5$ for both backbone architectures. Moreover, the ResNet architecture seems more robust to $\alpha$ than the VGG architecture, although both architectures offer a similar level of gains with the proper setting of $\alpha$. 
Most importantly, the consistently positive performance gains across all values of $\alpha$ suggest that \textit{any kind of gradient balancing} (even in suboptimal regimes) is healthy for multitask network training.

\subsection{Additional Experiments on a Multitask Facial Landmark Dataset}

\begin{figure}[htb!]
	\centering
	\includegraphics[width=0.5\textwidth]{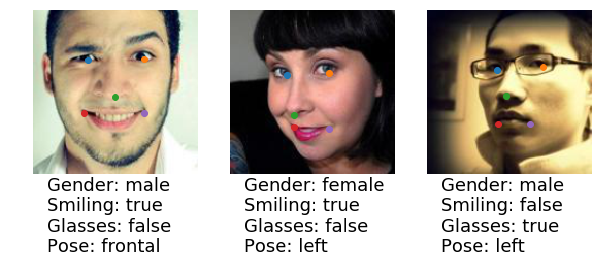}
	\caption{\textbf{Examples from the Multi-Task Facial Landmark (MTFL) dataset}. }
	\label{fig:mtfl_viz}
\end{figure}

We perform additional experiments on the Multitask Facial Landmark (MTFL) dataset \cite{zhang2014facial}. This dataset contains approximately 13k images of faces, split into a training set of 10k and a test set of 3k. Images are each labeled with $(x,y)$ coordinates of five facial landmarks (left eye, right eye, nose, left lip, and right lip), along with four class labels (gender, smiling, glasses, and pose). Examples of images and labels from the dataset are given in Figure \ref{fig:mtfl_viz}. 

The MTFL dataset provides a good opportunity to test GradNorm, as it is a rich mixture of classification and regression tasks. We perform experiments at two different input resolutions: 40x40 and 160x160. For our 40x40 experiments we use the same architecture as in \cite{zhang2014facial} to ensure a fair comparison, while for our 160x160 experiments we use a deeper version of the architecture in \cite{zhang2014facial}; the deeper model layer stack is [CONV-5-16][POOL-2][CONV-3-32]$^2$[POOL-2][CONV-3-64]$^2$[POOL-2][[CONV-3-128]$^2$[POOL-2]]$^2$[CONV-3-128]$^2$[FC-100][FC-18], where CONV-X-F denotes a convolution with filter size X and F output filters, POOL-2 denotes a 2x2 pooling layer with stride 2, and FC-X is a dense layer with X outputs. All networks output 18 values: 10 coordinates for facial landmarks, and 4 pairs of 2 softmax scores for each classifier.

\begin{table*}[htb!]
\caption{\textbf{Test error on the Multi-Task Facial Landmark (MTFL) dataset for GradNorm and various baselines}. Lower values are better and best performance for each task is bolded. Experiments are performed for two different input resolutions, 40x40 and 160x160. In all cases, GradNorm shows superior performance, especially on gender and smiles classification. GradNorm also matches the performance of \cite{zhang2014facial} on keypoint prediction at 40x40 resolution, even though the latter only tries to optimize keypoint accuracy (sacrificing classification accuracy in the process). }\vspace{0.2cm}
\footnotesize
\centering
\begin{tabular}{c c c c c c c c}
\hline
 & \textbf{Input}  & \textbf{Keypoint}  & \textbf{Failure.} & \textbf{Gender} & \textbf{Smiles} & \textbf{Glasses} & \textbf{Pose}\\
\textbf{Method} & \textbf{Resolution}&\textbf{Err. (\%)}&\textbf{Rate. (\%)}&\textbf{Err. (\%)}&\textbf{Err. (\%)}&\textbf{Err. (\%)}&\textbf{Err. (\%)}\\
\hline
Equal Weights&40x40 & 8.3 & 27.4 & 20.3 & 19.2 & 8.1 & 38.9\\
\cite{zhang2014facial}&40x40 & 8.2 & \textbf{25.0} & - & - & - & -\\
\cite{kendallarxiv}&40x40 & 8.3 & 27.2 & 20.7 & 18.5 & 8.1 & 38.9 \\
GradNorm $\alpha=0.3$& 40x40& \textbf{8.0} & \textbf{25.0} & \textbf{17.3} & \textbf{16.9} & 8.1 & 38.9\\
\hline
Equal Weights&160x160 & 6.8 & 15.2 & 18.6 & 17.4 & 8.1 & 38.9\\
\cite{kendallarxiv}&160x160 & 7.2 & 18.3 & 38.1 & 18.4 & 8.1 & 38.9\\
GradNorm $\alpha=0.2$& 160x160& \textbf{6.5} & \textbf{14.3} & \textbf{14.4} & \textbf{15.4}& 8.1 & 38.9\\
\end{tabular}


\label{table:testaccface}
\end{table*}

The results on the MTFL dataset are shown in Table \ref{table:testaccface}. Keypoint error is a mean over L$_2$ distance errors for all five facial landmarks, normalized to the inter-ocular distance, while failure rate is the percent of images for which keypoint error is over 10\%. For both resolutions, GradNorm outperforms other methods on all tasks (save for glasses and pose prediction, both of which always quickly converge to the majority classifier and refuse to train further). GradNorm also matches the performance of \cite{zhang2014facial} on keypoints, even though the latter did not try to optimize for classifier performance and only stressed keypoint accuracy. It should be noted that the keypoint prediction and failure rate improvements are likely within error bars; a 1\% absolute improvement in keypoint error represents a very fine sub-pixel improvement, and thus may not represent a statistically significant gain. Ultimately, we interpret these results as showing that GradNorm significantly improves classification accuracy on gender and smiles, while at least matching all other methods on all other tasks.

We reiterate that both glasses and pose classification always converge to the majority classifier. Such tasks which become ``stuck'' during training pose a problem for GradNorm, as the GradNorm algorithm would tend to continuously increase the loss weights for these tasks. For future work, we are looking into ways to alleviate this issue, by detecting pathological tasks online and removing them from the GradNorm update equation.  

Despite such obstacles, GradNorm still provides superior performance on this dataset and it is instructive to examine why. After all loss weights are initialized to $w_i(0) = 1$, we find that \cite{kendallarxiv} tends to increase the loss weight for keypoints relative to that of the classifier losses, while GradNorm aggressively decreases the relative keypoint loss weights. For GradNorm training runs, we often find that $w_{\text{kpt}}(t)$ converges to a value $\leq 0.01$, showing that even with gradients that are smaller by two orders of magnitude compared to \cite{kendallarxiv} or the equal weights method, the keypoint task trains properly with no attenuation of accuracy. 

To summarize, GradNorm is the only method that correctly identifies that the classification tasks in the MTFL dataset are relatively undertrained and need to be boosted. In contrast,  \cite{kendallarxiv} makes the inverse decision by placing more relative focus on keypoint regression, and often performs quite poorly on classification (especially for higher resolution inputs). These experiments thus highlight GradNorm's ability to identify and benefit tasks which require more attention during training.
\end{document}